\documentclass[11pt]{article}

\usepackage[preprint]{acl}

\usepackage{times}
\usepackage{latexsym}
\usepackage[LGR,T1]{fontenc}
\usepackage[utf8]{inputenc}
\usepackage{textgreek}
\DeclareTextFontCommand{\gk}{\fontencoding{LGR}\selectfont}
\usepackage{microtype}
\usepackage{inconsolata}
\usepackage{graphicx}
\usepackage{float}
\usepackage{booktabs}
\usepackage{tabularx}
\usepackage{multirow}
\usepackage{amsmath}
\usepackage{amssymb}
\usepackage{nicefrac}
\usepackage{xcolor}
\usepackage{pifont}
\usepackage{enumitem}
\usepackage{tikz}
\usetikzlibrary{positioning, arrows.meta, fit, backgrounds, calc, shapes.geometric}

\graphicspath{{media/}}

\newcommand{\result}[1]{#1}

\definecolor{prefixcol}{HTML}{0F766E}
\definecolor{suffixcol}{HTML}{3B5BA5}
\definecolor{middlecol}{HTML}{B4472E}
\definecolor{hintcol}{HTML}{C08A2D}
\definecolor{modelcol}{HTML}{334155}
\definecolor{beamcol}{HTML}{64748B}
\definecolor{bglight}{HTML}{F8FAFC}
\definecolor{bgtint}{HTML}{F1F5F9}
\definecolor{bordergrey}{HTML}{CBD5E1}
\newcommand{\fimchip}[3]{{\setlength{\fboxsep}{2.5pt}%
  \colorbox{#1}{\textcolor{#2}{\ttfamily\footnotesize #3}}}}

\title{Apollo Restore: A Foundation LLM for Historical Greek Optimized for
Fill-in-the-Middle Restoration of Ancient Greek Texts}

\author{Hope McGovern\textsuperscript{1*}\quad
        Anna Dolganov\textsuperscript{2*}\quad
        Samuel Belkadi\textsuperscript{1}\quad\\
        \textbf{Guillaume Kunsch\textsuperscript{1}}\quad
        \textbf{Dimitris Vlitas\textsuperscript{3}}\quad 
        \textbf{David A. Smith\textsuperscript{4}} \\
        \textsuperscript{1}Mistral AI \quad
        \textsuperscript{2}Austrian Academy of Sciences\quad
        \textsuperscript{3}Reply\quad
        \textsuperscript{4}Northeastern University\\ 
         \small{
         * equal contribution; these authors share first authorship
         }\\ 
        \small{
            \textbf{Correspondence:} \href{mailto:hope.speirs@mistral.ai}{hope.speirs@mistral.ai},
            \href{mailto:Anna.Dolganov@oeaw.ac.at}{Anna.Dolganov@oeaw.ac.at},
            \href{mailto:dasmith@ccs.neu.edu}{dasmith@ccs.neu.edu}
        }\\ \\
        \Large{\textsc{Preprint: Do Not Distribute}}}

\begin{document}
\maketitle
\begin{abstract}
We present Apollo Restore, a 24-billion-parameter large language model for restoring
\emph{lacunae}---physical gaps---in fragmentary Ancient Greek texts. Fine-tuned
from Mistral Small with a fill-in-the-middle objective, Apollo Restore reconstructs
missing spans without requiring oracle knowledge of their length. To our
knowledge, it is the first large-scale decoder model for historical Greek, and the
first for any ancient Mediterranean language.
Evaluated as in prior work, on short gaps of up to ten characters, Apollo Restore places
the correct restoration among its top twenty candidates for
\result{80.6\%}\,/\,\result{54.6\%}\,/\,\result{61.0\%} of documentary-papyrus,
literary-papyrus, and stone-inscription lacunae, exceeding the strongest
published models by \result{$1.6\times$}\,/\,\result{$2.6\times$}\,/\,%
\result{$1.4\times$}.
Prior evaluation
protocols, however, inflate scores through a bias toward trivially short gaps;
under a length-balanced metric Apollo Restore's advantage over the strongest published
models grows to \result{$2.3\times$}\,/\,\result{$3.5\times$}\,/\,%
\result{$1.6\times$} and degrades gracefully, even given incorrect length
hints. In a blind study, \result{20} expert papyrologists, epigraphists, and philologists strongly preferred Apollo Restore to the strongest baseline and judged its performance \textit{at least as good} as human restorations in \result{77\%} of cases.
Apollo Restore also improves the published reading of P.Herc.~1667---a papyrus roll carbonised in the eruption of Vesuvius in 79 \textsc{ce} and digitally unrolled and edited after Apollo Restore's training data was compiled. Apollo Restore is an output of the \textit{Decoding Antiquity} initiative to build specialized LLMs for historical languages and manuscripts, led by the Austrian Academy of Sciences.

\end{abstract}

\begin{figure}[t]
\centering
\begin{tabular}{@{}c@{\hspace{4pt}}c@{}}
  \includegraphics[height=3.6cm]{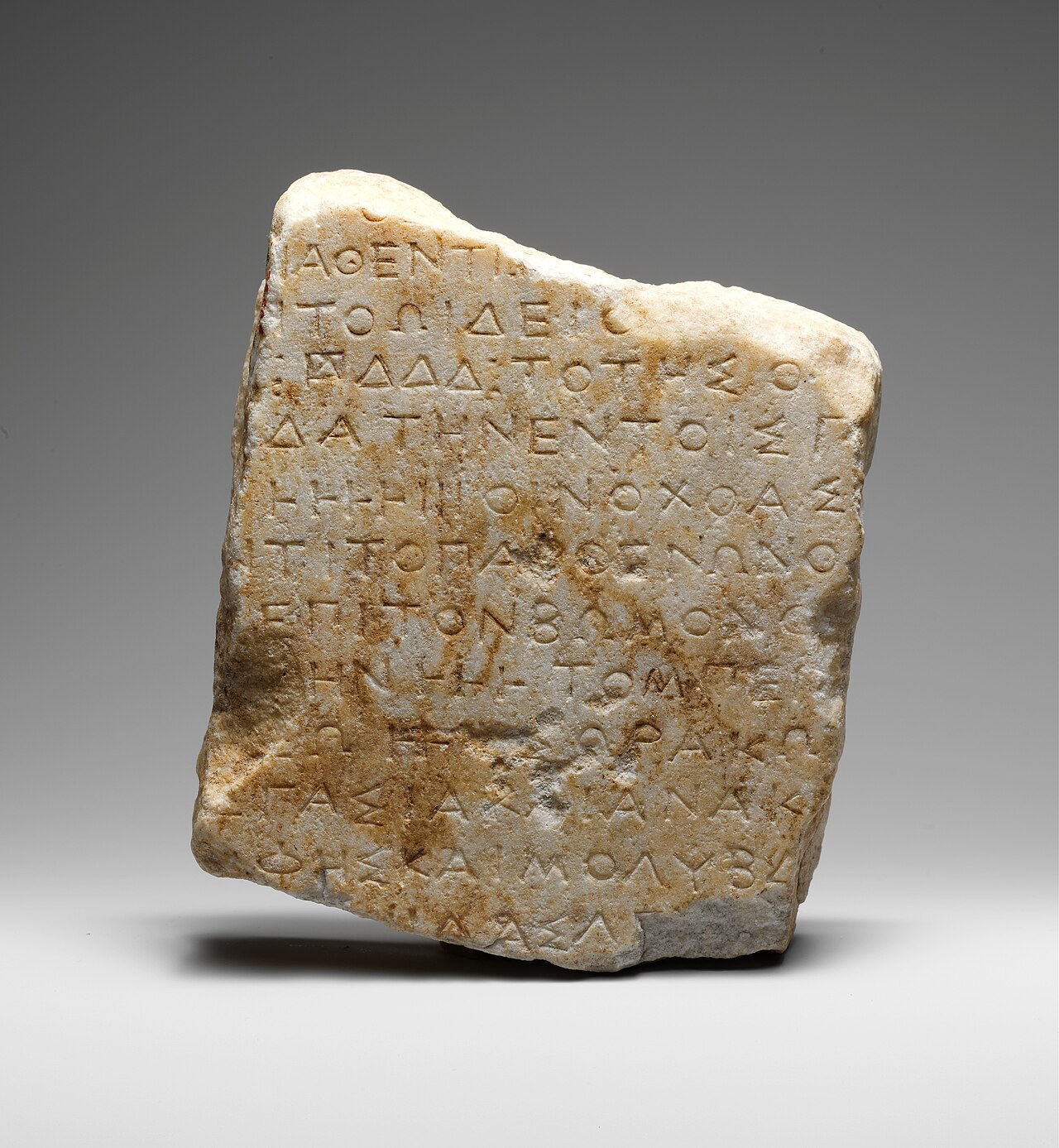} &
  \includegraphics[height=3.6cm]{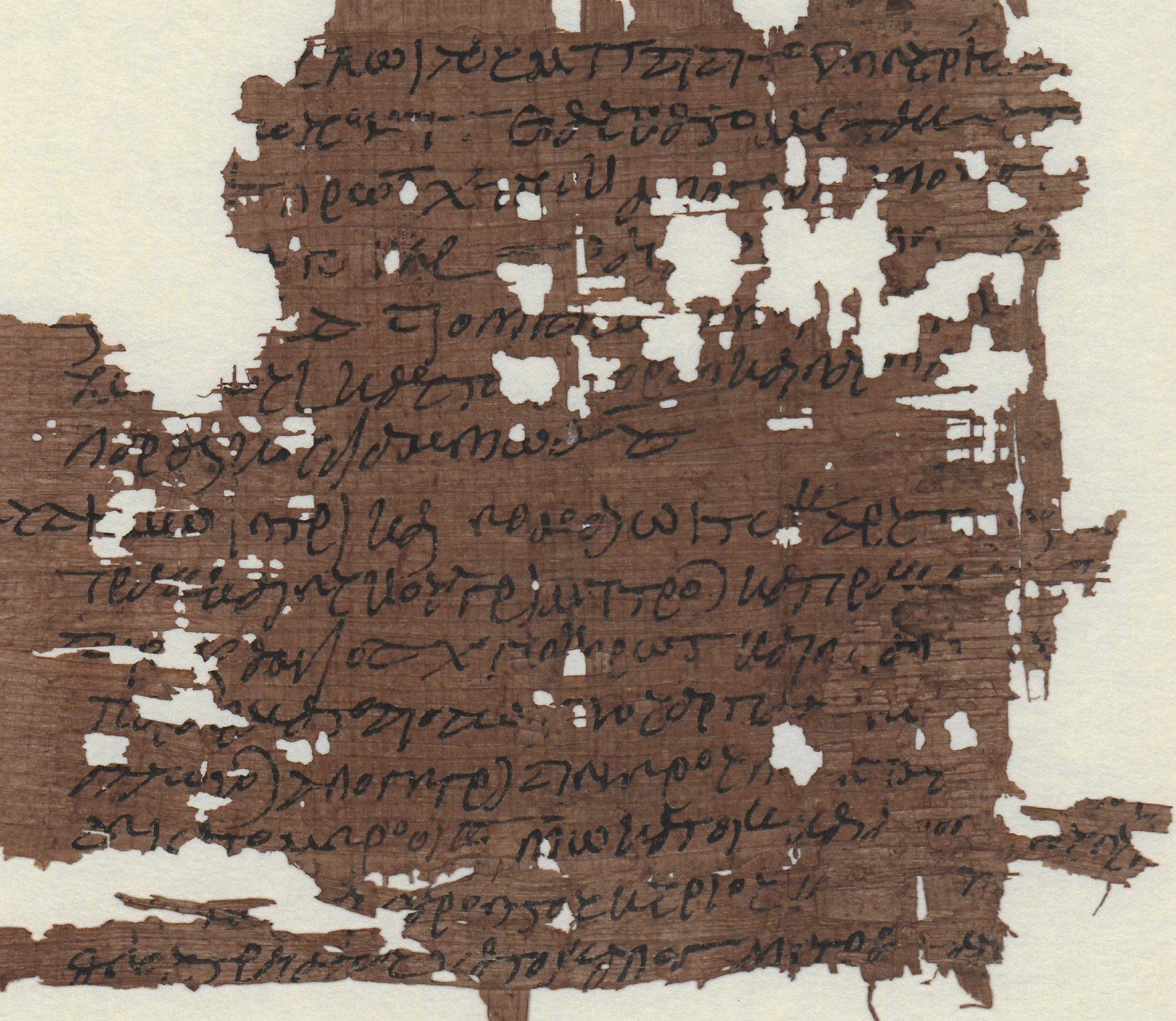} \\[2pt]
  {\footnotesize (a) stone inscription} &
  {\footnotesize (b) documentary papyrus} \\
\end{tabular}
\caption{Why length hints do not transfer from stone to papyrus or reflect field
conditions in epigraphy and papyrology.
\emph{(a)}~A marble inscription (IG~II\textsuperscript{2}~1688, Attica,
4th~c.\ \textsc{bce}), laid out on a regular grid whose fixed letter spacing lets
editors estimate missing characters---the assumption prior restoration models
rely on. \emph{(b)}~A documentary papyrus (CPR~XV~24, Roman Egypt,
119~\textsc{ce}), where elastic cursive writing makes a lacuna length difficult
to estimate. In both, gap length is impossible to gauge where the document is
broken off. Apollo Restore removes the dependency on a known gap length.}
\label{fig:artifacts}
\end{figure}

\begin{figure*}[t]
\centering
\begin{tikzpicture}[
    >=Stealth, font=\sffamily,
    panel/.style={draw=bordergrey, rounded corners=6pt, fill=white,
      line width=0.7pt, inner sep=8pt},
    hdr/.style={font=\footnotesize\bfseries\sffamily\color{modelcol}},
    chevron/.style={-{Stealth[length=4mm,width=4mm]}, line width=1.4pt,
      color=modelcol!55},
  ]
  \node[panel] (p1) {%
    \begin{tabular}{c}
      \includegraphics[width=4.0cm]{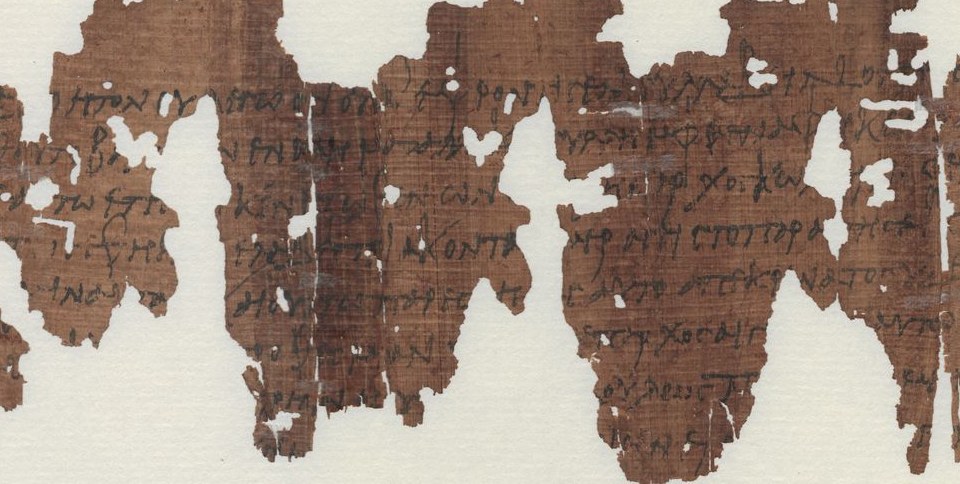}\\[3pt]
      {\scriptsize\ttfamily
        \textcolor{prefixcol}{...$\alpha\upsilon\tau$o$\hat{\upsilon}$}%
        \colorbox{middlecol!85}{\textcolor{white}{\,?\,?\,?\,}}%
        \textcolor{suffixcol}{$\tau\hat{\eta}\varsigma$...}}%
    \end{tabular}};
  \node[hdr, above=1pt of p1.north west, anchor=south west] {(1)\ Damaged artifact};
  \node[panel, right=1.15cm of p1, align=center] (p2) {%
    \begin{tabular}{@{}c@{}}
      \fimchip{suffixcol!14}{suffixcol}{[SUFFIX]}\;%
      \fimchip{suffixcol!7}{suffixcol}{$\tau\hat{\eta}\varsigma$...}\\[5pt]
      \fimchip{prefixcol!14}{prefixcol}{[PREFIX]}\;%
      \fimchip{prefixcol!7}{prefixcol}{...$\alpha\upsilon\tau$o$\hat{\upsilon}$}\\[5pt]
      \fimchip{hintcol!22}{hintcol}{[5-8]}\;%
      \fimchip{middlecol!22}{middlecol}{[MIDDLE]}\\[7pt]
      {\scriptsize\color{gray}hint: \texttt{[7]} exact\,/\,\texttt{[5-8]} range\,/\,none}
    \end{tabular}};
  \node[hdr, above=1pt of p2.north west, anchor=south west] {(2)\ FIM input (reordered)};
  \node[panel, right=1.15cm of p2, align=center] (p3) {%
    \begin{tabular}{@{}c@{}}
      {\setlength{\fboxsep}{5pt}\colorbox{modelcol}{\textcolor{white}{%
        \sffamily\small\bfseries Apollo Restore \;\mdseries\scriptsize 24B}}}\\[6pt]
      {\scriptsize\color{gray}beam search $k{=}20$}\\[4pt]
      {\small\ttfamily
        \begin{tabular}{@{}r@{\;}l@{\;\;}l@{}}
          \scriptsize1.&\textcolor{middlecol}{$\mu\eta\nu\grave{o}\varsigma$}&\scriptsize\color{gray}0.82\\
          \scriptsize2.&\textcolor{middlecol}{$\mu\eta\nu\acute{o}\varsigma$}&\scriptsize\color{gray}0.09\\
          \scriptsize3.&\textcolor{middlecol}{$\mu\eta\nu\hat{\omega}\nu$}&\scriptsize\color{gray}0.04\\
        \end{tabular}}
    \end{tabular}};
  \node[hdr, above=1pt of p3.north west, anchor=south west] {(3)\ Ranked restorations};
  \draw[chevron] (p1.east) -- (p2.west);
  \draw[chevron] (p2.east) -- (p3.west);
\end{tikzpicture}
\caption{Overview of the Apollo Restore pipeline. (1)~Transcription of a damaged artifact with physical
lacunae; the text around each gap is decomposed into a
\textcolor{prefixcol}{prefix} and \textcolor{suffixcol}{suffix} bracketing the
\textcolor{middlecol}{missing middle}. (2)~These are encoded as a
fill-in-the-middle sequence in suffix--prefix--middle order with an optional
\textcolor{hintcol}{length hint}. (3)~Apollo Restore generates ranked restorations by
beam search.}
\label{fig:model-overview}
\end{figure*}
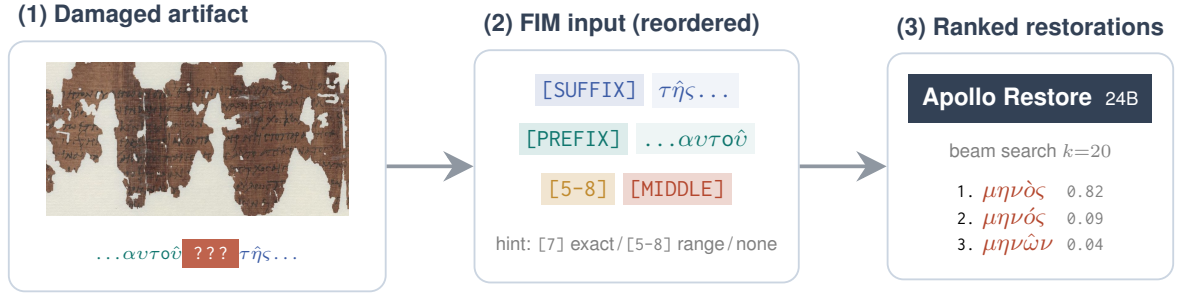

\section{Introduction}
\label{sec:introduction}

More than one million papyri have been discovered around the Mediterranean and
the Middle East, the majority written in Ancient Greek during the
``papyrological millennium'' between the conquests of Alexander and the rise of
the Islamic caliphates (ca.\ 300~\textsc{bce}--700~\textsc{ce}). Surviving in a
fragile, fragmentary state, most papyri are documentary remnants of ordinary
life---receipts, letters, legal records, census
data~\citep{palmeRangeDocumentaryPapyri2009, bagnallEverydayWriting2011}---that
offer unparalleled insight into the ancient world. Reconstructing the
\emph{lacunae}---the physical gaps where the writing surface has perished---demands
hundreds of hours from specialists. After a century and a half of scholarship,
less than ten percent of surviving papyri have been
published (papyri.info)~leaving nearly a million
unread documents \citep{vanMinnenMillenniumPapyrology2007}. An automated system proposing plausible reconstructions would be
an invaluable instrument for accelerating the decipherment of antiquity.

Recent work has applied neural architectures to ancient text
restoration~\citep{assaelRestoringAttributingAncient2022,
assaelContextualizingAncientTexts2025,
cullhedInstructTuningPretrainedCausal2024}. This work, however, rests on an assumption that is defensible in \emph{epigraphy} (the study of stone inscriptions) but largely untenable in \emph{papyrology}: that the lengths of lacunae are known (Figure~\ref{fig:artifacts}). Inscriptions are laid out on a regular grid, allowing scholars to count missing letters. But papyrus column widths may vary within a single text~\citep{dolganovForgeryFiscalFraud2023} and scribes can compress ten cursive characters into a void that would visually accommodate four. In both media, texts are often broken on one or more sides, rendering length estimates hypothetical
or impossible. Restoration models that require length hints are thus inapplicable
to a large proportion of real research cases.

We further observe that, because short gaps are routinely restored by editors,
existing models severely overfit to small lacunae ($\leq 3$ characters). While
this mirrors the training distribution, it diminishes practical utility: short
gaps are exactly the cases domain experts resolve without algorithmic help. As we
show in \S\ref{sec:evaluation}, this overfitting, combined with
short-gap-dominated test sets, \emph{inflates the reported accuracy of prior
systems}, masking their collapse in accuracy on longer, truly difficult gaps.

Finally, prior systems are limited by narrow, task-specific training: Ithaca
\citep{assaelRestoringAttributingAncient2022} trains only on a subset of Greek
inscriptions---a tiny fraction of digitized Greek---while
\citet{cullhedInstructTuningPretrainedCausal2024} instruction-tune an
open-weight LLM with no additional pretraining on Greek.

Figure~\ref{fig:model-overview} summarises our approach to the problem of text restoration.  Our contributions are:

\begin{enumerate}[nosep,leftmargin=*]
  \item \textbf{Apollo Restore, the first LLM for an ancient Mediterranean
    language.} Built on Mistral Small and trained on ca.\ 600M words of Greek
    from antiquity to 1900---the largest open corpus of digitized Greek to
    date---Apollo Restore is, to our knowledge, the first large-scale decoder model for
    historical Greek. 
  \item \textbf{A length-hint-free restoration method.} We reframe restoration as
    a variable-length fill-in-the-middle task, so Apollo Restore infers gap length from
    context and remains usable when lacuna size is unknown.
  \item \textbf{A balanced evaluation framework.} We show that prior metrics
    inflate scores via short-gap bias, and introduce uniform-weighted, length-
    robust metrics. The advantage of Apollo Restore over prior work \emph{grows} on the harder,
    more realistic long-gap regime (up to \result{$3.5\times$}).
  \item \textbf{Expert-validated quality.} In a blind study with \result{20}
    professional scholars, the restorations of Apollo Restore were preferred to the strongest
    baseline's in a large majority of judgements.
  \item \textbf{The first AI-assisted restoration of a Herculaneum papyrus.} We apply Apollo Restore to P.Herc.~1667, a philosophical text carbonised by the
    eruption of Vesuvius in 79~\textsc{ce} and recovered through virtual unrolling only in 2026
    \citep{angelotti2026completevirtualunwrappingreading}, and propose new restorations to a text unreadable
    for centuries (Appendix~\ref{sec:appendix-qualitative-herc}).

\end{enumerate}

\section{Related Work}
\label{sec:related-work}

We first discuss the two most directly comparable projects on restoring lacunae in Greek texts and then survey other related work.

\paragraph{Ithaca.}
Building on the earlier Pythia character-level model for Greek epigraphy
\citep{assaelRestoringAncientText2019}, \citet{assaelRestoringAttributingAncient2022}
introduced Ithaca, a multitask BigBird~\citep{zaheer2020bigbird} Transformer that
jointly restores, geographically attributes, and dates Greek inscriptions,
trained on a subset of I.PHI (78{,}608 inscriptions from the Packard Humanities Institute);
the approach was recently extended to multimodal inputs in Aeneas
\citep{assaelContextualizingAncientTexts2025}. Ithaca has advanced epigraphic restoration but has three
limitations for general use. First, it requires an \emph{exact} one-to-one
mapping of placeholder to missing characters; its no-count mode
(the \texttt{\#} symbol) sharply degrades quality because it was not designed as a
variable-length generator. Second, its input is restricted to 50--750 characters
and its UI caps restorations at 20 characters. Third, training only on datable
inscriptions---a tiny fraction of digitized Greek---leaves it unable to restore
non-epigraphic text.

\paragraph{Cullhed.}
\citet{cullhedInstructTuningPretrainedCausal2024} instruction-tunes Llama~3.1~8B
\citep{grattafiori2024llama3} for papyrus restoration, using GPT-4o-mini for data augmentation, and omit spaces
to mirror \emph{scriptio continua}. However, space omission elides the variable
word-spacing of much documentary and even literary
material~\citep{turnerGreekManuscripts1971}, and the model still conditions on an
explicit character-count hint in its prompt---preserving the length-oracle
dependency that limits real-world use. Neither system performs additional
pretraining on a broad historical-Greek corpus, and neither integrates retrieval
into generation.

\paragraph{Other work.} The Logion project trained masked language models \citep[MLMs;][]{devlin2019bert} on premodern Greek to flag and correct likely scribal corruptions in transmitted texts \cite{cowen-breen-etal-2023-logion, brooks-etal-2025-annotated}. Its objective is error \emph{correction} rather than lacuna restoration, and its word-level MLM conditions on both sides of a single token; it does not generate the multi-character missing spans that Apollo Restore targets.  In discussing similar experiments for Latin, \citet{bamman2020latinbertcontextuallanguage} observe that, while their MLM is trained on synthetic gaps, it is important to evaluate on more realistic restoration tasks.  Their experiments restore single words of at least two characters, marked by textual editors with angle brackets as conjectures or from a parallel source, and in sentences with 10--100 words of context.  Working with pretrained MLMs for restoring lacunae in Latin manuscripts, \citet{zhang-etal-2026-lt4hala} address the mismatch between a model's tokenization and the placement and length of gaps, and \citet{locaputoFillingLacunaeAncient} likewise study lacuna-filling in Latin inscriptions.  More broadly, computational study of historical languages is an active area \citep{sommerschieldMachineLearningAncient2023}, with recent work probing general-purpose LLMs for classical philology \citep{riemenschneiderExploringLargeLanguage2023} and historical-language tasks such as part-of-speech tagging \citep{stussiPartofSpeechTagging16thCentury2024} and intertextual analysis \citep{gong-etal-2025-augmented}.  Apollo Restore differs from all of the above in coupling large-scale continued pre-training on a broad historical-Greek corpus with a variable-length fill-in-the-middle objective for multi-character restoration.

\section{Method}
\label{sec:method}

\subsection{Task Formulation}
\label{subsec:task-formulation}

We frame restoration as a \textbf{fill-in-the-middle} (FIM) task for an
autoregressive LM \cite{bavarianEfficientTrainingLanguage2022, fried2023incodergenerativemodelcode}.
A contiguous span is masked; the model receives the text
before (prefix) and after (suffix) the gap and generates the missing middle.
Each example is packed in suffix ($s$)--prefix ($p$)--middle ($m$) order,
\begin{center}\small
\texttt{[BOS][SUFFIX]}\,$s$\,\texttt{[PREFIX]}\,$p$\,%
$\langle h \rangle$\,\texttt{[MIDDLE]}\,$m$\,\texttt{[EOS]}
\end{center}
where \texttt{[PREFIX]}, \texttt{[SUFFIX]}, and \texttt{[MIDDLE]} are pre-existing
reserved tokens of the base tokenizer and
$\langle h \rangle$ is an optional length hint
(\S\ref{subsec:length-hints}). Training uses standard causal-LM
cross-entropy, but loss is computed \emph{only over the middle span and its
terminal} \texttt{[EOS]}; control and hint tokens are loss-masked. FIM is thus
achieved by input reordering and masking alone, not a bespoke objective. This
gives two advantages over instruction-tuning: the model learns gap
length \emph{implicitly} from context rather than from an oracle count, and the
suffix--prefix--middle structure mirrors how a papyrologist reads
around a lacuna.

\subsection{Data}
\label{subsec:data}

The training data of Apollo Restore comprises three corpora of primary-document transcriptions (documentary papyri, DDB; literary papyri, DCLP; inscriptions, I.PHI) and the Corpus of Open Greek (COG), a compilation of public-domain and open-access editions (Table~\ref{tab:data-sources}). Primary-document transcriptions follow the Leiden
Conventions~\citep{van_groningen_signis_1932}, where brackets mark lacunae with
proposed restorations. Documents are segmented into windows of up to 500
characters ($\pm 30\%$, breaking at spaces) and formatted as FIM triples. Each
fragmentary source holds out 5\% of documents for validation and 5\% for test.
COG aggregates ca.\ 600M words from the Open Greek and Latin project, the
Institutional Books 1.0 corpus~\citep{cargnelutti_institutional_2025}, and the
\emph{Patrologia Graeca}; about 69\% predates 1453. Full corpus-construction
details, dating, and deduplication are given in Appendix~\ref{sec:appendix-data}.

\begin{table}[t]
\centering\small
\begin{tabular}{@{}lp{5.5cm}@{}}
\toprule
\textbf{Source} & \textbf{Domain} \\
\midrule
DDB  & Documentary papyri (Duke Databank) \\
DCLP & Literary papyri \\
I.PHI   & Stone inscriptions \cite{assaelRestoringAncientText2019} \\
COG  & Printed editions (Corpus of Open Greek) \\
\bottomrule
\end{tabular}
\caption{Training data sources.}
\label{tab:data-sources}
\end{table}

\paragraph{Text normalisation.}
All text is lowercased, NFD-decomposed with combining diacritics stripped (final
sigma preserved; iota subscript expanded to adscript to retain morphology).
Digits are removed, punctuation collapses to a middle-dot, and the hyphen is
reserved as a gap marker. The Leiden apparatus is resolved consistently
(Appendix~\ref{sec:appendix-data}).

\paragraph{Inscriptions: train on I.PHI, evaluate on IG.}
We train inscriptions on I.PHI (Ithaca's source) but \emph{evaluate} on the openly redistributable Inscriptiones Graecae (IG), which also measures cross-collection generalisation. Importantly, we train on the \emph{entire} I.PHI collection---all \result{$\sim$156{,}000} inscriptions with surviving text---rather than the datable subset of \result{78{,}608} used by Ithaca \citep{assaelRestoringAttributingAncient2022}. Ithaca restricts training to inscriptions carrying geographic and chronological labels because it jointly predicts region and date; Apollo Restore performs only restoration and therefore imposes no such requirement, roughly doubling the available epigraphic training data. Date and place metadata are retained where present but are never used to filter the corpus.  For evaluation on IG, we drop any inscription that shares an exact $25$-character Greek run with the I.PHI train or test data, then remove non-linguistic or unreadable fragments. 
This yields \result{1{,}482} provably novel inscriptions, so the epigraphic results measure genuine cross-collection generalisation rather than memorisation.

\subsection{Synthetic Gap Construction}
\label{subsec:synthetic-gaps}

We distinguish \emph{restored gaps} (text reconstructed by scholars) from
\emph{true gaps} (irrecoverably lost). Rather than train on the small, length-biased set
of scholarly restorations, we in-fill all existing restorations to obtain a clean
base text and then synthetically mask new spans---so \textbf{the model never sees
scholarly restorations during training}. This gives \emph{scale} (millions of
examples vs.\ tens of thousands) and a corrected \emph{distribution}: scholarly
restorations are overwhelmingly 1--3 characters, whereas we sample gap lengths
uniformly on $[1,25]$, ensuring signal for the longer, more challenging lacunae.
Gaps are sampled on the fly (FIM ratio $1.0$, windows $\geq 50$ characters);
$10\%$ use an empty suffix and $10\%$ an empty middle. Scholarly restorations are
reserved for validation and test.

\subsection{Uncertain Readings and Fair Comparison}
\label{subsec:uncertain-context}

Editions distinguish a \emph{lacuna} (lost material) from an \emph{uncertain
reading} (a partially legible character difficult to identify, dotted in Leiden). Targets
are always genuinely secure spans. For the surrounding context we adopt the
\textbf{uncertain-included} setting (uncertain characters treated as certain) as
the headline condition, because this matches how both baselines preprocess text
\citep{assaelRestoringAttributingAncient2022,
cullhedInstructTuningPretrainedCausal2024}; the stricter uncertain-excluded
setting is retained as an ablation (Appendix~\ref{sec:appendix-uncertain}).

\subsection{Length-Hint Conditioning}
\label{subsec:length-hints}

Apollo Restore does not require a length hint but can exploit one. During training, hints
are given as a mix of \textbf{exact} (40\%, e.g.\ \texttt{[7]}), an \textbf{estimated range}
(40\%, e.g.\ \texttt{[5-8]}), or \textbf{absent} (20\%).
Crucially, 15\% of hints are \emph{deliberately incorrect}, preventing the model
from treating hints as hard constraints and teaching graceful degradation. Hint
tokens are loss-masked. At inference, real fragments with multiple gaps are filled
left-to-right, each prediction feeding the next
(Appendix~\ref{sec:appendix-multigap}).

\section{Evaluation Framework}
\label{sec:evaluation}

\subsection{Metrics and the short-gap bias}
\label{subsec:metrics}

We report Top-1 accuracy, Top-20 accuracy (correct answer among 20 beam
candidates), character accuracy, and length delta $\Delta_{\mathrm{len}}$
(predicted minus true gap length). All models are scored with the same
\emph{space-agnostic} match (removing spaces, punctuation, and numerals) that
Cullhed and Ithaca use, so comparisons are like-for-like.

\paragraph{Uniform-weighted metrics.}
The labelled test data is heavily skewed toward short lacunae: gaps of 1--3
characters make up roughly 69\% of samples. A naive mean therefore rewards a
model mostly for the easy short gaps that experts already restore unaided. We
compute \textbf{uniform-weighted} metrics: group samples by exact gap length,
average within each length, then average lengths with equal weight---so a bucket
with 500 examples counts the same as one with 5. This reflects the full range of
archaeologically realistic gap sizes and is a better proxy for real utility.

\paragraph{Prior scores are inflated by this bias.}
The correction is substantial and strengthens our claims. On documentary papyri, evaluated as prior work does over short gaps of 1--10 characters (Table~\ref{tab:main-results}, prior protocol), the Top-1 advantage of Apollo Restore over the strongest baseline is \result{$2.0\times$} (\result{0.603} vs.\ \result{0.302}); weighting gap lengths equally over the full 1--20 range (our protocol) widens it to \result{$2.8\times$} (\result{0.432} vs.\ \result{0.155}). On literary papyri the same shift grows the Top-20 margin from \result{$2.6\times$} to \result{$3.5\times$}. The baselines' apparent competitiveness is largely an artefact of a test distribution dominated by easy short gaps; on the longer gaps that matter, they collapse (\S\ref{subsec:per-gap}). 

\subsection{Baselines}
\label{subsec:baselines}

We compare against the two strongest published systems,
\textbf{Ithaca}~\citep{assaelRestoringAttributingAncient2022} (inscriptions) and
\textbf{Cullhed}~\citep{cullhedInstructTuningPretrainedCausal2024} (papyri),
using their released models unmodified, each in its native input format, on the
identical held-out gaps, with 20 beam candidates. Both are given the true length
hint (Cullhed requires one; Ithaca fills exactly its placeholder count and cannot
run hint-free). We preserve each model's own preprocessing so it sees text as its
authors intended. Table~\ref{tab:model-comparison} contrasts the three systems;
Apollo Restore is alone in supporting flexible (or absent) length hints and full
space-agnostic restoration across all genres of historical Greek.

\begin{table}[t]
\centering\small
\begin{tabular}{@{}lccc@{}}
\toprule
\textbf{Feature} & \textbf{Ithaca} & \textbf{Cullhed} & \textbf{Apollo Restore} \\
\midrule
Size            & custom     & 8B        & \textbf{24B} \\
Paradigm        & multitask  & instruct  & \textbf{FIM} \\
Length hint     & exact only & exact only & \textbf{flexible} \\
Hint-free mode  & \ding{55}  & \ding{55} & \textbf{\ding{51}} \\
Space-agnostic  & \ding{55}  & \ding{51} & \textbf{\ding{51}} \\
Non-epigraphic  & \ding{55}  & \ding{51} & \textbf{\ding{51}} \\
\bottomrule
\end{tabular}
\caption{Feature comparison. Only Apollo Restore works with no/approximate length hints and all genres of Greek.}
\label{tab:model-comparison}
\end{table}

\section{Results}
\label{sec:results}

Apollo Restore is fine-tuned from Mistral Small Base 2503 (24B; \citealp{jiang2023mistral7b}) for 10{,}000 steps; the
full training configuration and infrastructure are given in
Appendix~\ref{sec:appendix-repro}.

\subsection{Main Results}
\label{subsec:main-results}

Table~\ref{tab:main-results} reports results under two protocols side by side.
Under the \emph{prior-work} protocol (an unweighted sample mean over short gaps
of $1$--$10$ characters), Apollo Restore reaches Top-20 accuracy of \result{80.6\%}
(DDB), \result{54.6\%} (DCLP), and \result{61.0\%} (IG) already \result{$1.6\times$}\,/\,\result{$2.6\times$}\,/\,%
\result{$1.4\times$} the strongest baseline. Our \emph{uniform-weighted} metric
over the full $1$--$20$ range (\S\ref{subsec:metrics}) corrects the
short-gap bias and \emph{widens} the relative advantage of Apollo Restore to
\result{$2.3\times$}\,/\,\result{$3.5\times$}\,/\,\result{$1.6\times$} in Top-20.
Apollo Restore also tracks the true gap length closely ($\Delta_{\mathrm{len}}$ near
zero) while the baselines systematically over-generate.

\begin{table}[t]
\centering\small
\setlength{\tabcolsep}{3pt}
\begin{tabular}{@{}llcccccc@{}}
\toprule
& & \multicolumn{2}{c}{\textbf{Prior}} & \multicolumn{3}{c}{\textbf{Our protocol}} \\
\cmidrule(lr){3-4}\cmidrule(lr){5-7}
\textbf{Corpus} & \textbf{Model} & \textbf{T1} & \textbf{T20} & \textbf{W.\,T1} & \textbf{W.\,T20} & $\boldsymbol{\Delta_{\mathrm{len}}}$ \\
\midrule
\multirow{2}{*}{DDB}  & \textbf{Apollo Restore} & \textbf{\result{0.603}} & \textbf{\result{0.806}} & \textbf{\result{0.432}} & \textbf{\result{0.632}} & \result{$-$0.6} \\
                      & Cullhed         & \result{0.302} & \result{0.490} & \result{0.155} & \result{0.272} & \result{$+$1.5} \\
\midrule
\multirow{2}{*}{DCLP} & \textbf{Apollo Restore} & \textbf{\result{0.297}} & \textbf{\result{0.546}} & \textbf{\result{0.175}} & \textbf{\result{0.327}} & \result{$-$2.2} \\
                      & Cullhed         & \result{0.108} & \result{0.208} & \result{0.048} & \result{0.094} & \result{$+$1.5} \\
\midrule
\multirow{2}{*}{IG}   & \textbf{Apollo Restore} & \textbf{\result{0.403}} & \textbf{\result{0.610}} & \textbf{\result{0.248}} & \textbf{\result{0.393}} & \result{$+$0.3} \\
                      & Ithaca          & \result{0.312} & \result{0.430} & \result{0.178} & \result{0.241} & \result{$-$1.0} \\
\bottomrule
\end{tabular}
\caption{Main results (space-agnostic, uncertain-included, exact hint).
\textbf{Prior protocol:} unweighted sample-mean Top-1/Top-20 over short gaps of
$1$--$10$ characters---the T20 figures
(\result{80.6\%}/\result{54.6\%}/\result{61.0\%}) quoted in the abstract.
\textbf{Our protocol:} uniform-weighted Top-1/Top-20 (W.\,T1/W.\,T20) over the
full $1$--$20$ range, which corrects the short-gap bias and \emph{widens} the
relative advantage of Apollo Restore. Ithaca is evaluated on IG (open) though trained on I.PHI.}
\label{tab:main-results}
\end{table}

\subsection{Per-Gap-Length Analysis}
\label{subsec:per-gap}

\begin{figure}[t]
\centering
\includegraphics[width=\columnwidth]{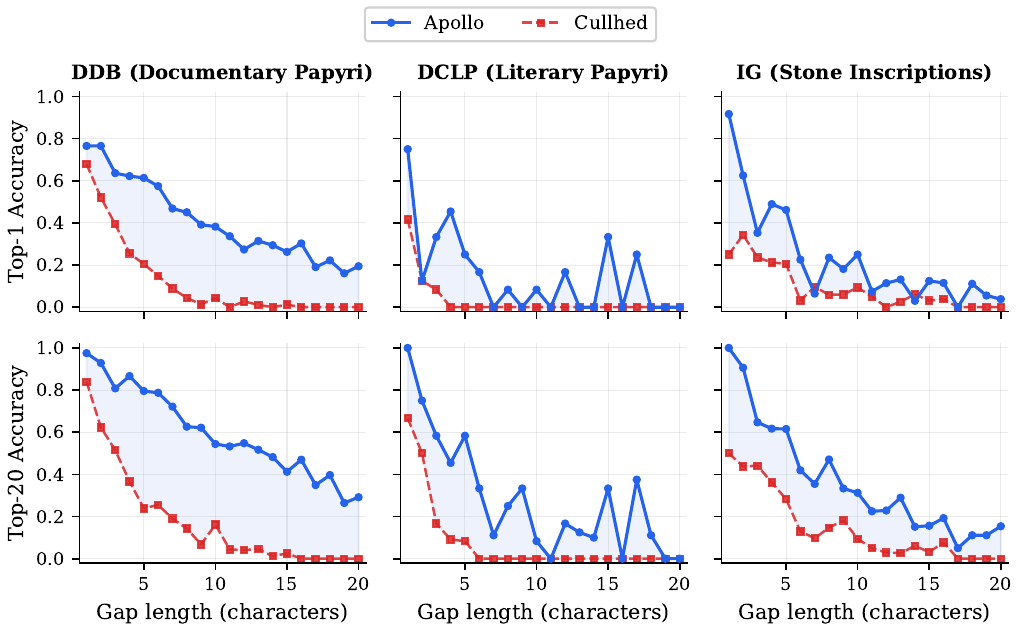}
\caption{Per-gap-length Top-1 (upper) and Top-20 (lower) accuracy, Apollo Restore vs.\
the strongest baseline per corpus. Baseline accuracy collapses on longer gaps;
Apollo Restore degrades gracefully.}
\label{fig:per-gap}
\end{figure}

Figure~\ref{fig:per-gap} resolves accuracy by gap length and is the clearest evidence for our argument about metrics. Apollo Restore maintains strong accuracy across the full $1$--$20$ range, whereas the baselines are competitive only on the shortest gaps and collapse toward zero beyond roughly $8$--$14$ characters---precisely the regime where algorithmic help is most valuable.


\subsection{Length-Hint Robustness}
\label{subsec:hint-robustness}

Real gap lengths are rarely known with precision. Table~\ref{tab:hint-robustness} shows Apollo Restore degrades gracefully as hints worsen: its weighted Top-1 falls only from \result{0.432} (exact) to \result{0.364} (fully wrong) and \result{0.332} (no hint), retaining roughly three-quarters of its accuracy with \emph{no} length information---while beating Cullhed's exact-hint score by more than 2$\times$. Cullhed
cannot operate without a hint. Training with deliberately incorrect hints
(\S\ref{subsec:length-hints}) enables this robustness.

\begin{table}[t]
\centering\small
\begin{tabular}{@{}lcc@{}}
\toprule
\textbf{Hint condition} & \textbf{Apollo Restore W.\,T1} & \textbf{Cullhed W.\,T1} \\
\midrule
Exact          & \result{0.432} & \result{0.155} \\
10\% wrong     & \result{0.428} & \result{0.150} \\
50\% wrong     & \result{0.398} & \result{0.139} \\
100\% wrong    & \result{0.364} & \result{0.128} \\
No hint        & \result{0.332} & --- \\
\bottomrule
\end{tabular}
\caption{Length-hint robustness on the full DDB test set. Each hint
condition re-evaluates the same gaps under a different hint quality. Apollo Restore with
\emph{no} hint (\result{0.332}) still exceeds Cullhed's exact-hint score by more
than $2\times$. ``---'': Cullhed has no hint-free mode.}
\label{tab:hint-robustness}
\end{table}

\subsection{Comparison with Frontier Closed Models}
\label{subsec:closed-models}

We further compare against three frontier general-purpose models via
OpenRouter\footnote{Accessed \result{2025} via \url{https://openrouter.ai}; model strings \texttt{openai/gpt-5.1}, \texttt{openai/gpt-5-mini}, \texttt{anthropic/claude-opus-4.8}.}
on a fixed \result{200}-gap DDB sub-sample (Top-1, space-agnostic; identical
sub-sample re-scored for all systems). Apollo Restore far exceeds every closed model
(Table~\ref{tab:closed-models}): the best, Claude Opus~4.8 (\result{0.151}),
merely matches the specialised Cullhed baseline and trails Apollo Restore by
\result{$3.0\times$}. The flagships respect the length hint and recover short
gaps, so this is a genuine capability ceiling, not a formatting artefact. Strikingly, \emph{Apollo Restore with no hint}
beats every closed model given an \emph{exact} hint.

\begin{table}[t]
\centering\small
\begin{tabular}{@{}llcc@{}}
\toprule
\textbf{Model} & \textbf{Type} & \textbf{Exact} & \textbf{No hint} \\
\midrule
\textbf{Apollo Restore}& specialised FIM (24B)     & \textbf{\result{0.447}} & \result{0.366} \\
Cullhed          & specialised instr.\ (8B)  & \result{0.147} & --- \\
Claude Opus~4.8  & frontier flagship         & \result{0.151} & \result{0.102} \\
GPT-5.1          & frontier flagship         & \result{0.103} & \result{0.091} \\
GPT-5-mini       & frontier efficiency       & \result{0.005} & \result{0.000} \\
\bottomrule
\end{tabular}
\caption{Frontier closed models on DDB (Top-1 uniform-weighted over $1$--$20$,
space-agnostic, \result{200}-gap sub-sample). Apollo Restore with no hint exceeds every
closed model given an exact hint.}
\label{tab:closed-models}
\end{table}

\subsection{Agreement with Scholarly Restorations on Real Lacunae}
\label{subsec:silver-agreement}

The results above use \emph{synthetic} gaps, where the masked characters are known
with certainty. We additionally evaluate on \emph{true lacunae}: real gaps a
modern editor has restored by conjecture. Here the reference is a scholarly
\emph{silver} \textit{label}, not ground truth, and a divergent model reading may still be
philologically defensible; we therefore report \emph{agreement} with the
published restoration---agree@1 (top-1 matches the editor), agree@20 (the editor's
reading appears among the 20 beams), and mean character error rate
(CER; normalised Levenshtein distance, \citealp{levenshtein1966binary}) against
the editor's reading. Table~\ref{tab:silver-agreement} shows Apollo Restore agrees with
published scholarship more often than every baseline on every corpus, and by a
wide margin on inscriptions (agree@1 \result{0.478} vs.\ Ithaca's \result{0.194}).
This is a conservative test---Apollo Restore is penalised whenever it proposes a defensible
alternative to the editor---yet it leads throughout, confirming that the
synthetic-gap gains transfer to genuine restoration.

\begin{table}[t]
\centering\small
\begin{tabular}{@{}llccc@{}}
\toprule
\textbf{Corpus} & \textbf{Model} & \textbf{Ag@1} & \textbf{Ag@20} & \textbf{CER}$\downarrow$ \\
\midrule
\multirow{2}{*}{DDB}  & \textbf{Apollo Restore} & \textbf{\result{0.630}} & \textbf{\result{0.859}} & \textbf{\result{0.375}} \\
                      & Cullhed         & \result{0.555} & \result{0.770} & \result{0.382} \\
\midrule
\multirow{2}{*}{DCLP} & \textbf{Apollo Restore} & \textbf{\result{0.426}} & \textbf{\result{0.737}} & \textbf{\result{0.544}} \\
                      & Cullhed         & \result{0.324} & \result{0.640} & \result{0.627} \\
\midrule
\multirow{2}{*}{IG}   & \textbf{Apollo Restore} & \textbf{\result{0.478}} & \textbf{\result{0.747}} & \textbf{\result{0.621}} \\
                      & Ithaca          & \result{0.194} & \result{0.416} & \result{0.724} \\
\bottomrule
\end{tabular}
\caption{Agreement with scholarly (silver-label) restorations on \emph{true
lacunae}, exact hint, space-agnostic. Ag@1/Ag@20: editor's reading matches the
top-1 / appears in the top-20. Apollo Restore leads on every corpus; higher agreement and
lower CER are better.}
\label{tab:silver-agreement}
\end{table}

\subsection{Language-Modelling Quality}
\label{subsec:perplexity}

\begin{figure}[t]
\centering
\includegraphics[width=\columnwidth]{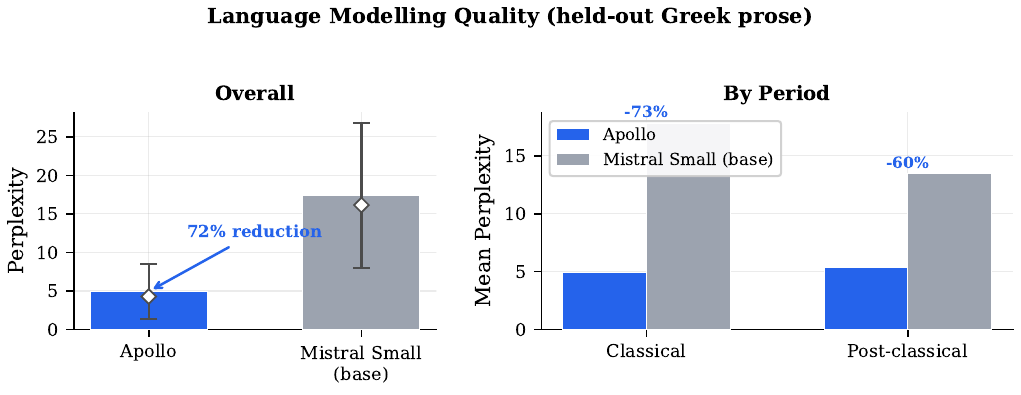}
\caption{Language-modelling quality on held-out Greek prose. Left: overall
perplexity (mean $\pm$ std; diamond = median). Right: by time period.
Apollo Restore reduces perplexity $3.5\times$ over the base model.}
\label{fig:perplexity}
\end{figure}

Beyond gap-filling, we ask whether FIM fine-tuning improves Apollo Restore's general modelling of historical Greek. Figure~\ref{fig:perplexity} reports perplexity on held-out Greek prose against the base Mistral Small model. Apollo Restore achieves \result{$3.5\times$} lower perplexity (\result{4.94} vs.\ \result{17.37}), with consistent gains across chronological periods (\result{$-72.6\%$} classical, \result{$-60.2\%$} post-classical). FIM fine-tuning thus does not merely teach gap-filling: it substantially sharpens the model's token-level command of historical Greek.


\section{Human Evaluation}
\label{sec:human-eval}

Automatic metrics on synthetic gaps cannot tell us whether the restorations of Apollo Restore are \emph{plausible for scholars}. We therefore ran a blind preference study and a timed restoration experiment with \result{20} professional papyrologists, epigraphists, and philologists, matched to the three corpora, on \emph{real} lacunae.

\subsection{Part~I: Blind Preference Study}
\label{subsec:parti}

Each expert completed a questionnaire in two phases. In \textbf{Phase~A} (model vs.\ model), a lacuna in context was shown with the deduplicated, shuffled union
of Apollo Restore's and the domain baseline's top-3 readings behind neutral labels; the
expert marked each plausible reading and picked the best. In \textbf{Phase~B} (model
vs.\ human), Apollo Restore's top-1 was shown blind against the editor's published
restoration, restricted to cases where the two \emph{disagreed}, asking whether
Apollo Restore's alternative is competitive with expert judgement. The returned data
comprise \result{696} Phase~A and \result{874} Phase~B judgements.

\paragraph{Apollo Restore is strongly preferred (Phase~A).}
Across every corpus, experts chose Apollo Restore's reading far more often than the
baseline's (Table~\ref{tab:humaneval}; Figure~\ref{fig:heA} in
Appendix~\ref{sec:appendix-qual}). On documentary
papyri Apollo Restore was judged best in \result{59.0\%} of items vs.\ Cullhed's
\result{13.8\%}---an \result{81.0\%} head-to-head win rate excluding ties. The
margin is largest exactly where prior systems fail: on \emph{long} gaps the
tie-excluded win rate rises to \result{88.1\%}, and with \emph{no} length hint
(the realistic setting) to \result{93.4\%}.

\paragraph{Apollo Restore sometimes reads better than the human expert (Phase~B).}
Phase~B is a deliberately adversarial test: it contains \emph{only} the cases
where Apollo Restore's top-1 \emph{disagreed} with the published editor---every case an
automatic metric counts as an outright error. Remarkably, even on this
worst-case subset, experts judged Apollo Restore's reading \emph{at least as good as} the
editor's published restoration in \result{36.7\%} of documentary-papyrus cases,
and \emph{strictly better} than the human expert in \result{16.0\%}
(Table~\ref{tab:phaseB}). That a model proposes a reading that a professional
papyrologist prefers to an established scholarly reading, in one in six of its
apparent ``mistakes,'' is a striking result that showcases Apollo Restore's potential to advance editorial work on historical Greek texts.

Because Phase~B excludes the majority of cases where Apollo Restore already
\emph{agrees} with the editor (Table~\ref{tab:phaseB}), \result{36.7\%} is a
strict lower bound: Apollo Restore's reading on all DDB lacunae is at least as good as
the editor's \result{77\%} of the time. Inter-annotator agreement on
shared anchor items is substantial for Phase~A (\result{$\kappa=0.67$} on DDB).

\begin{table}[t]
\centering\small
\begin{tabular}{@{}llcccc@{}}
\toprule
& & \textbf{Apollo} & \textbf{Tie} & \textbf{Base} & \textbf{Win\%}$^{\dagger}$ \\
\midrule
\multirow{3}{*}{\shortstack[l]{DDB\\{\scriptsize vs Cullhed}}}
  & overall   & \result{59.0} & \result{27.2} & \result{13.8} & \textbf{\result{81.0}} \\
  & long gaps & \result{54.9} & \result{37.7} & \result{7.4}  & \textbf{\result{88.1}} \\
  & no hint   & \result{73.4} & \result{21.4} & \result{5.2}  & \textbf{\result{93.4}} \\
\midrule
\multirow{2}{*}{\shortstack[l]{IG\\{\scriptsize vs Ithaca}}}
  & overall   & \result{57.6} & \result{14.1} & \result{28.3} & \textbf{\result{67.1}} \\
  & long gaps & \result{54.5} & \result{25.0} & \result{20.5} & \textbf{\result{72.7}} \\
\bottomrule
\end{tabular}
\caption{Phase~A blind model-vs-model preference (percent of judgements; DDB
vs.\ Cullhed, IG vs.\ Ithaca). $^{\dagger}$Win\% excludes ties,
$=\text{Apollo Restore}/(\text{Apollo Restore}+\text{Base})$. Apollo Restore is preferred across the
board, most strongly on long gaps and when no length hint is available.}
\label{tab:humaneval}
\end{table}

\begin{table}[t]
\centering\small
\begin{tabular}{@{}lcccc@{}}
\toprule
\textbf{Corpus} & \textbf{Apollo Restore} & \textbf{Equal} & \textbf{Human} & \textbf{Ap.\,$\geq$\,H.} \\
\midrule
DDB  & \result{16.0} & \result{20.7} & \result{55.2} & \textbf{\result{36.7}} \\
IG   & \result{19.8} & \result{16.8} & \result{56.3} & \textbf{\result{36.5}} \\
\bottomrule
\end{tabular}
\caption{Phase~B blind model-vs-human preference, restricted to the
\emph{disagreement} cases where Apollo Restore's top-1 differs from the editor (percent).
``Ap.\,$\geq$\,H.'' = Apollo Restore judged better or equal. Because the far larger set of
\emph{agreement} cases (agree@1$=\result{0.630}$ on DDB) is excluded, these figures are a strict \textbf{lower bound}: across \emph{all}
lacunae Apollo Restore's reading is at least as good as the editor's far more often
($\approx\result{77\%}$ on DDB). Even on this adversarial subset, Apollo Restore is judged
strictly \emph{better} than the human expert on \result{16.0\%} of DDB cases.
(DCLP omitted: single annotator.)}
\label{tab:phaseB}
\end{table}

\begin{figure}[t]
\centering
\includegraphics[width=0.82\columnwidth]{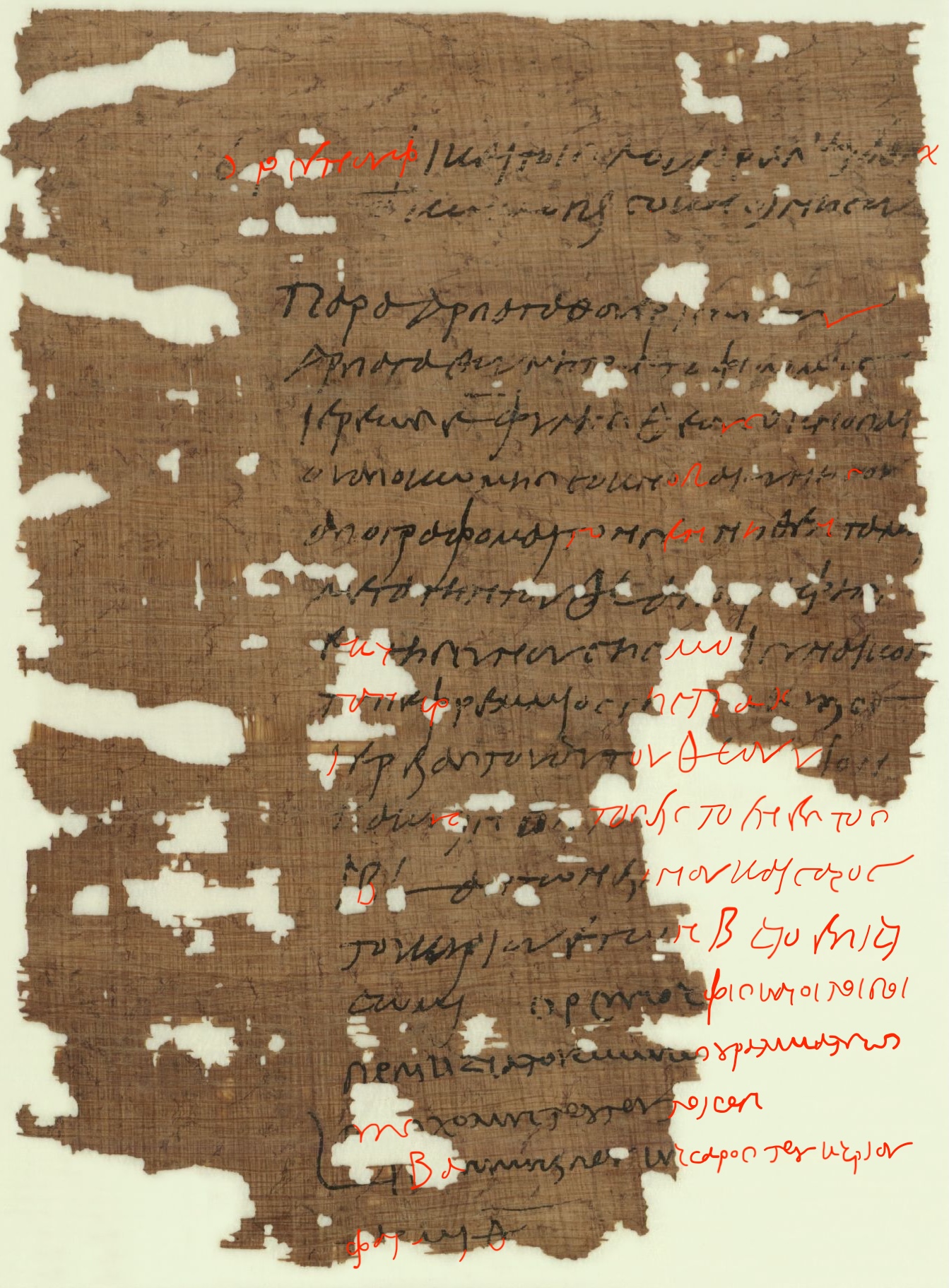}
\caption{Apollo Restore restoring a real papyrus: SPP~XXII~18, a birth declaration from
Soknopaiou Nesos (149~\textsc{ce}). Surviving ink is black; Apollo Restore's proposed
restorations are overlaid in red by a
papyrologist. Even without length hints, the model reconstructs missing spans of varying length directly
from the surrounding documentary context. Apollo Restore's top-1 restorations match published restorations and add one new, plausible restoration of \gk{grammat'ewc} in line 16.}
\label{fig:qualitative}
\end{figure}

\paragraph{Qualitative examples.}
Figure~\ref{fig:qualitative} shows Apollo Restore's restorations (in red) of a real documentary papyrus. Restorations in Table~\ref{tab:examples} and Appendix~\ref{sec:appendix-examples} show Apollo Restore recovering the editor's reading (typically as its top-2 candidate) and offering a divergent top-1
proposal that scholars judged equally plausible to the published one.

\begin{table}[t]
\centering\small
\setlength{\tabcolsep}{4pt}
\renewcommand{\arraystretch}{1.25}
\begin{tabularx}{\columnwidth}{@{}lX@{}}
\toprule
\multicolumn{2}{@{}l}{\textbf{SB~1~5337} --- documentary papyrus (account)} \\
\midrule
Editor & \gk{>Af[rod'iths]} \\
Apollo Restore & \gk{>Af[rod'iths p'olews]} \, (top-1, plausible) \newline
         \gk{>Af[rod'iths]} \, (top-2, = editor) \\
\addlinespace[3pt]
\midrule
\multicolumn{2}{@{}l}{\textbf{P.Herc.~1667} --- literary papyrus (philosophical text)} \\
\midrule
Editor & \gk{poie\~in <ws pef'ukamen} \newline \emph{\footnotesize ``to do as is in our nature''} \\
Apollo Restore & \gk{poik'ilws pef'ukamen} \, (top-1, plausible) \newline
         \emph{\footnotesize ``we are by nature diverse''} \newline
         \gk{poie\~in <ws pef'ukamen} \, (top-2, = editor) \\
\bottomrule
\end{tabularx}
\caption{Worked restoration examples. Apollo Restore recovers the editor's reading as a
high-ranked candidate while its top-1 offers a scholarly-plausible alternative. In
the literary example (P.Herc. 1667) Apollo Restore helps identify a textual problem in the published edition (Appendix~\ref{sec:appendix-qualitative-herc}).}
\label{tab:examples}
\end{table}

\subsection{Part~II: Timed Restoration Experiment}
\label{subsec:partii}

Part~II measured Apollo Restore's effect on the workflow itself: papyrologists and
epigraphists restored documentary excerpts with multiple lacunae, half unaided
(with database search) and half with Apollo Restore, counterbalanced for a within-item
comparison. Although these experts were already fast unaided (median
\result{3.0}~min), with Apollo Restore the within-item mean fell by \result{1.3}~min/text
(faster on \result{11}/\result{20} texts). Experts also recovered
several published restorations with Apollo Restore that they had missed without it
(Appendix~\ref{sec:appendix-qual}). Apollo Restore's value thus lies less in raw speed on
routine gaps than in assistance on harder cases. Experts praised its contextual
inference but noted inconsistent handling of formulaic expressions---a capability
targeted by the planned retrieval-augmented generation
\citep[RAG;][]{lewis2020rag} extension.

\section{Discussion}
\label{sec:discussion}

\paragraph{Why FIM beats instruction-tuning.}
Instruction-tuned models must parse a character count and emit exactly that number of
characters. FIM absorbs surrounding context and generates to
a natural stop, learning the context--length relationship implicitly. The
$\Delta_{\mathrm{len}}$ results confirm this: Apollo Restore tracks true gap length
closely (\result{$-0.2$} on DDB) while Cullhed over-generates
(\result{$+1.8$}).

\paragraph{Short-gap overfitting should not be rewarded.}
The gap between unweighted and weighted metrics exposes a pervasive problem in
prior evaluation, and we argue uniform-weighted metrics should become standard in
computational papyrology. Apollo Restore's top-20 contains the correct restoration
$\sim$60\% of the time on documentary papyri, enabling a shortlist-review
workflow rather than restoration from scratch.

\section{Conclusion}
\label{sec:conclusion}

We present Apollo Restore, the first large language model for an ancient
Mediterranean language, which reframes Ancient Greek restoration as
fill-in-the-middle and removes the oracle-length dependency of prior work. Apollo Restore
sets a new state of the art across documentary papyri, literary papyri, and
inscriptions, with its advantage \emph{growing} on the long, realistic gaps that
prior metrics obscured, and is preferred by scholars in blind
evaluation. We will release the weights, data, and evaluation code.

\section*{Limitations}

Apollo Restore is trained on digitised scholarly editions and inherits their biases:
geographic and temporal coverage is uneven, and texts with unusual orthography or
dialects remain challenging. Multi-lacuna inference is sequential and greedy, so
errors in early gaps can propagate to later ones. Our inscription evaluation uses
IG while training on I.PHI; while this measures cross-collection
generalisation, it prevents a strictly in-distribution epigraphic comparison. The
closed-model comparison is limited to Top-1 on a 200-gap DDB sub-sample for cost
reasons, and some reported figures are provisional pending a final re-run (marked
in the source). The human study, though blind and multi-annotator, involves
modest per-corpus annotator counts (especially a single literary-papyrus
respondent) and evaluates plausibility rather than ground-truth correctness.

\section*{Ethics Statement}

Apollo Restore restores fragmentary historical texts and proposes \emph{candidate}
readings for expert review; it is not a substitute for scholarly judgement, and
generated restorations should be labelled as machine-proposed. Training data
derives from openly licensed or public-domain corpora; access-restricted sources
(I.PHI) are used only for training and are neither redistributed nor used for the
released benchmark. We foresee no harm to human subjects; the expert study was
voluntary and anonymised.

\section*{Acknowledgments}

The authors would like to thank: our panel of expert scholars in papyrology, epigraphy, and philology for their help and valuable feedback; Alison Babeu, Giulia Taurino, and Cliff Wulfman for their help in compiling the Corpus of Open Greek; and Greg Leppert and the rest of Harvard's Institutional Data Initiative team for their work opening up their collection, including Greek books with improved OCR.  David Smith was supported in part by the Schmidt Sciences-funded project ``Beyond Translation: Opening up the Human Record.''

\bibliography{references,ancient_greek_fim}

\clearpage
\appendix

\section{Qualitative Analysis of Expert Evaluation}
\label{sec:appendix-qualitative}
\label{sec:appendix-qual}

\begin{figure}
\centering
\includegraphics[width=0.7\columnwidth]{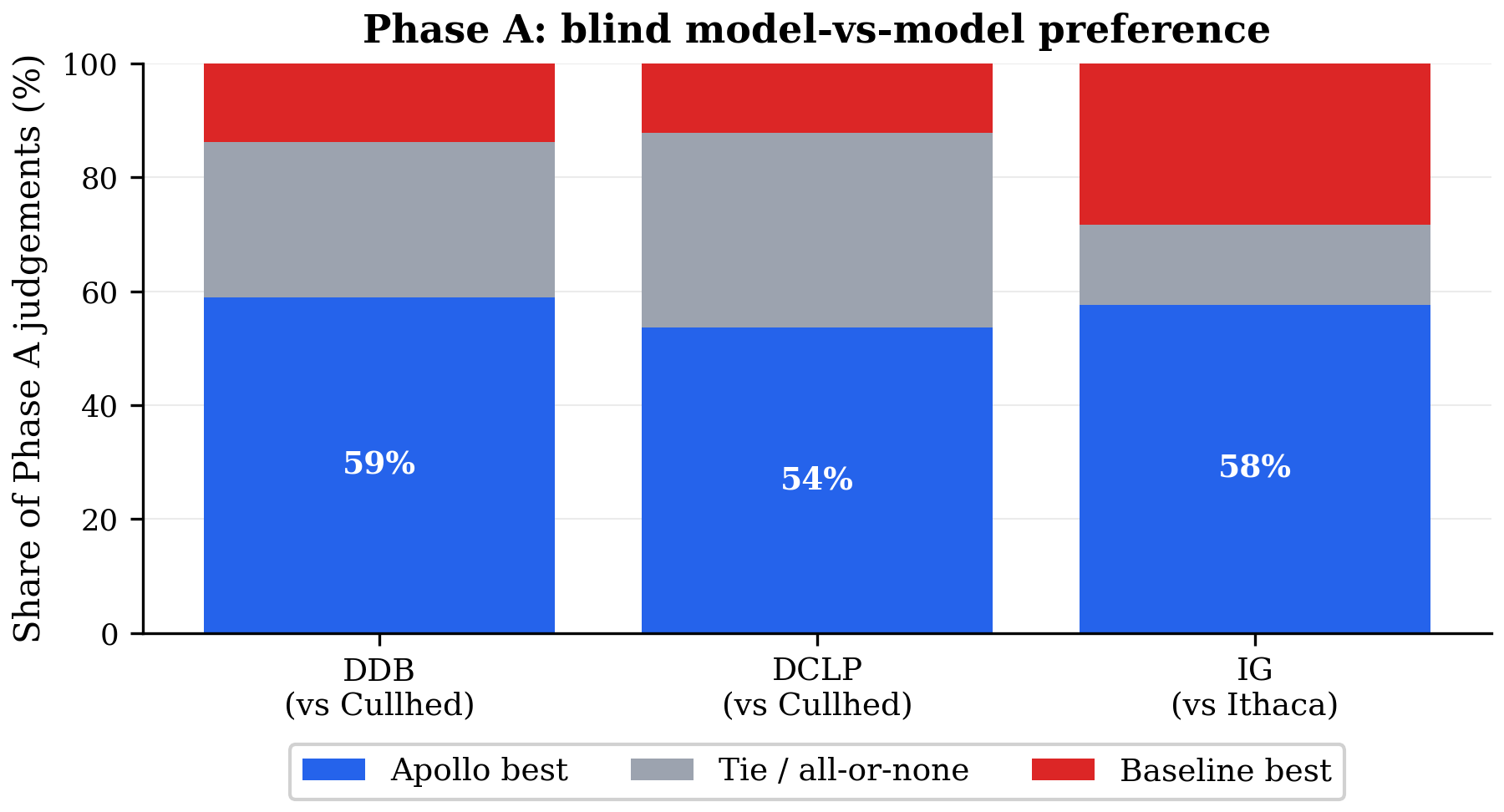}
\caption{Phase~A: share of blind judgements preferring each system's reading.
Ties include shared readings and ``all/none plausible'' verdicts. (Same data as
Table~\ref{tab:humaneval}.)}
\label{fig:heA}
\end{figure}

The evaluation of Apollo Restore was performed by experienced editors of inscriptions and papyri, including leading scholars who have published hundreds of texts. The evaluation therefore sets a high bar for critical scrutiny of the model. In addition to performing the structured evaluation (Part~I Blind Preference Study; Part~II Timed Restoration Experiment; Section~\ref{sec:human-eval}), evaluators were also asked to experiment with the model in their research and provide qualitative feedback. Figure~\ref{fig:heA} visualises the Phase~A preferences summarised in Table~\ref{tab:humaneval}.

\subsection{Impact on Text Restoration Workflows}
\label{sec:appendix-qualitative-impact}

In the independent portion of the timed experiment (involving the completion of multiple gaps in 10 texts, half with the support of Apollo Restore) evaluators were able to restore many lacunae quickly with a high level of accuracy. Apollo Restore enhanced restoration speed, lowering the within-item mean by 1.3 min/text, faster on 11/20 texts, which in view of the high level of editorial expertise among the evaluators means that Apollo Restore is expected to make further gains in accelerating restoration workflows within the broader research community.

Apollo Restore also had a significant positive impact on restoration quality. Evaluators working without Apollo Restore consistently missed a number of secure published restorations, which evaluators working with Apollo Restore successfully made. For example, in a tax receipt signed by a certain Germanos (O.Ber. II 137, Berenike, Roman Egypt, 65 \textsc{ce}) it is possible to restore Germanos as the missing name of the issuer. Of six evaluators given this text, three working without Apollo Restore failed to restore it, one restored it without Apollo Restore, and two restored it with the support of Apollo Restore, which produces the correct restoration as its top-1. 

As an another example, in an account listing multiple Egyptian villages in the late Roman province of Arcadia (SB I 5337 = Tyche 37 (2022) 111), the letters \gk{af} may be restored as \gk{>Af[rod\'ithc]} or \gk{>Af[rod\'ithc p\'olewc]}. Of five evaluators given this text, three working without Apollo Restore did not restore it, and two restored it with the support of Apollo Restore, which produces the correct restorations at ranks 1 and 2.

Of three epigraphic evaluators asked to restore \gk{t\~w| glukut\'atw|} [GAP]\gk{vw| <Rwman\~w|} in a Roman-period funerary inscription (IG X 2, 2, 1185, Carevi Kuli, 2--3c. \textsc{ce}), one evaluator working without Apollo Restore failed to restore it, and two evaluators working with Apollo Restore (which gives the correct restoration \gk{[t'ekn]w|} as its top-1) restored it.

It may be inferred that experts working under time pressure occasionally missed contextual clues and parallels supporting a restoration, but were quickly able to identify the correct restoration from a shortlist provided by Apollo Restore. The results also support the inference that researcher confidence in the feasibility of a restoration task improves when provided with a shortlist of possibilities from which to choose, rather than restoring from scratch. The feedback of one expert makes this point explicit: ``there were almost always suggestions that were very useful. A few times, however, I chose a different reconstruction over the one Apollo Restore had identified as the best.''

These results showcase the utility of Apollo Restore in supporting and accelerating text restoration workflows. Model capabilities to restore text are expected to be further strengthened by retrieval-augmented restoration and iterative training on user feedback.  In addition to improving restoration accuracy, we expect retrieval of parallel passages to strengthen scholarly confidence in the AI system.

\subsection{Other Qualitative Observations}
\label{sec:appendix-qualitative-observations}

Epigraphic evaluators remarked that the randomly chosen evaluation texts were highly fragmentary, with gaps that were often difficult to restore with confidence. Nevertheless, in the timed experiment two epigraphic evaluators explicitly assessed 73.33\% (11 of 15) of restorations by Apollo Restore as correct or plausible and 26.67\% (4 of 15) as incorrect or unlikely. Two epigraphic evaluators noted that Apollo Restore tended to favour restorations that would be more likely in papyri than inscriptions, such as restoring a name with Am---ades as Ammoniades, or interpreting \gk{str\~wsin} as bedding rather than street pavements.

Evaluators noted that providing no length hints occasionally produced better results, while in other cases the absence of length hints resulted in empty outputs where restorations fell below the probability threshold of the model. These observations will be tested and analysed in a separate publication. One epigraphic evaluator observed that results improved when targeting specific restorations for Apollo Restore to perform, leaving other gaps unrestored or providing confident restorations as text, thereby enhancing the context window seen by the model. 

Evaluators consistently made positive observations about Apollo Restore's capacity for contextual inference. For example, an epigraphic evaluator tested an inscription from Messene where Apollo Restore improves upon the editio princeps (SEG XLI 323, 1991) by correctly inferring in the lacuna two of three categories of competitions mentioned in the text, increasing the total estimated lacuna length and altering the restoration of the entire text. In this, Apollo Restore matches the observations of a scholarly re-edition of the inscription (SEG LXII 224, 2012). Other evaluators were impressed that in a fragmentary funerary epigram (IG IX 1², 4, 970, Corcyra, 2--1c. \textsc{bce}) Apollo Restore picked up on its metrical and archaizing nature and restored the epic form \gk{A>i[akid\~an]} at rank 1 and the iambic \gk{A>i[ak\'oc]} at rank 2 (both unlikely in this case, as one would expect the name of the deceased in the accusative at the end of the first line). However, Apollo Restore's contextual understanding also had its limits: one evaluator noted that Apollo Restore did not consistently restore as its top-1 the orthographical variants \gk{>indikti\~wnoc} and \gk{>indikti\'onoc} in line with the respective form present elsewhere in the text. This observation will be tested and analysed separately.

\subsection{Example Restoration of P.Herc. 1667}
\label{sec:appendix-qualitative-herc}

An interesting restoration experiment involved the new philosophical text recently recovered through digital unrolling of a carbonised papyrus from Herculaneum in the Bay of Naples, part of an ancient library preserved by the eruption of Vesuvius in 79 \textsc{ce} \cite{angelotti2026completevirtualunwrappingreading, fowlerHerculaneumPapyri2024}. Apollo Restore made numerous suggestive restorations that will be analysed in a separate publication. Here we focus on one example.

The recovered text speaks about the constitution of human nature and the role of reasoned judgment (\gk{fr\'onhsic}) in controlling human impulses. In column 11, lines 7--10, the editors give:
\begin{quote}
\gk{pr\d{`o}c [d]`e \d{t}\d{`a} o\'<u\d{t}\d{w}\d{c} \'<eka\textsigma{}}|\gk{ta {po}}\textbf{\ldots}\gk{{\d{w}\d{c}} pef\'uka}|\gk{men ka\`i pr\`oc t\`o ta\~u}|\gk{ta poie\~in `<a fa\'inein}\ldots

And for each of these things in this way we are by nature \ldots and for doing these things that seem \ldots
\end{quote}

\noindent Suggested editorial restorations are \gk{po\d{n}\d{h}\d{r}\d{\~w}\d{c} pef\'uka}|\gk{men} = ``we are by nature base/cowardly'' or \gk{poi\d{e}\d{\~i}\d{n} \d{<w}\d{c} pef\'uka}|\gk{men} = ``to do as is in our nature.'' Philologically, \gk{ponhr\~wc} seems unlikely, as \gk{ponhr\'oc} is attributive, and it is unclear why the circumlocution \gk{pef\'ukamen} with the adverbial form should be used. While \gk{poie\~in <wc} creates a parallel construction, the syntax of the restored \gk{pr`oc [d]`e t`a o\'<utwc \'<ekasta poie\~in}  would clash with the subsequent \gk{pr`oc t`o ta\~uta poie\~in}. The editors appear to be aware of the problem and suggest an alternative reading of \gk{t`o o\'<utwc} in the apparatus.

When given the text as printed, Apollo Restore suggests \gk{poik\'ilwc pef\'uka}|\gk{men} = ``we are by nature diverse'' as its top-1. The choice is appropriate: \gk{poik\'iloc} and its variants appear 44 times in 28 Herculaneum papyri, including 23 papyri of Philodemus, in whose writings the term typically occurs with the sense of ``varied, diverse'' and the adverb \gk{poik\'ilwc} as ``in various ways'' (see, e.g., P.Herc. 1674 col. 53, line 15 and col. 56, line 11). The supplement \gk{poik\'ilwc} would be plausible in the context and considering that the adjective \gk{poik\'iloc} is typically attributive for objects rather than persons the indirect expression \gk{poik\'ilwc pef\'ukamen} would be justified. The editors' second suggested restoration \gk{poie\~in <wc} is offered by Apollo Restore at rank 2.

However, when given the alternative reading \gk{pr`oc [d]`e t`o o\'<utwc \'<ekasta} mentioned by the editors in the apparatus, which resolves the syntactic irregularity of the passage, Apollo Restore returns \gk{poie\~in <wc}, which fits the parallel construction and constitutes philologically preferable Greek. Examination of the papyrus images included in the publication strongly supports \gk{pr`oc [d]`e t`o o\'<utwc \'<eka\textsigma{}}|\gk{ta poie\~in <wc pef\'uka}|\gk{men ka`i pr`oc t`o ta\~u}|\gk{ta poie\~in `<a fa\'inein} as the most probable reading. Thus, in addition to making restorations grounded in the relevant textual corpus, Apollo Restore supports editors in correcting problems in their text. 

This example further highlights Apollo Restore's capacity for contextual inference and ability to support text restoration tasks across all genres of historical Greek, including the extraordinary philological challenges posed by the complex fragmentary corpus of Hellenistic philosophy preserved in the Herculaneum papyri. 

\subsection{General Impressions of Evaluators}
\label{sec:appendix-qualitative-impressions}

Evaluators were consistently positive in their comments and looked forward to the future development and expansion of the model, notably through the addition of user feedback and a search system to enable retrieval-augmented restoration and retrieval of comparable passages across the Greek textual tradition. General comments include the following: 
\begin{itemize}
    \item ``I was surprised by the quality of the results.''
    \item ``For the most part, Apollo Restore was very helpful.''
    \item ``The fact that the reconstruction of formulas is to be learned automatically in the future would be a major step forward.''
    \item ``There was hardly any text that could be supplemented beyond a few letters, but Apollo Restore did its job well and swiftly.''
    \item ``Apollo Restore is already working quite well! There were almost always suggestions that were very useful.''
    \item ``I want to congratulate you on taking on this project, which will be extremely useful for the entire community.''
\end{itemize}

\section{Worked Restoration Examples}
\label{sec:appendix-examples}

Table~\ref{tab:examples-appendix} gives the full context, editorial reading, and Apollo Restore's top candidates for one example per genre, including the stone inscription omitted from the main text. In every case, Apollo Restore recovers the editor's reading as its rank-2 candidate while offering a scholarly-plausible rank-1 alternative.  In the literary example, its top-1 was judged plausible and points to problems in the published text, as discussed in Appendix~\ref{sec:appendix-qualitative-herc}. Examples were selected and annotated by a domain expert; a hyphen
in brackets marks the restored span.

\begin{table*}[t]
\centering\small
\setlength{\tabcolsep}{5pt}
\renewcommand{\arraystretch}{1.3}
\begin{tabularx}{\textwidth}{@{}lX@{}}
\toprule
\multicolumn{2}{@{}l}{\textbf{Documentary papyrus} --- SB~1~5337 (account, Late Roman Egypt), lines 5--6} \\
\midrule
Context & \gk{>ap`o Ne'ilou p'olews (ka`i) >Af[rod'iths ...]} $\mid$ \gk{<o(mo\~u) l'i(trai)} \dots\ (``from Neilopolis and Aphroditopolis \dots\ total 1,506 pounds'') \\
Editor  & \gk{>Af[rod'iths]} \\
Apollo Restore  & \gk{>Af[rod'iths p'olews]} (top-1, plausible); \quad \gk{>Af[rod'iths]} (top-2, matches editor) \\
\addlinespace[2pt]
\midrule
\multicolumn{2}{@{}l}{\textbf{Literary papyrus} --- P.Herc.~1667 (philosophical text), col.~11 lines 7--10} \\
\midrule
Context & \gk{pr`os [d]`e t`a o<'utws <'ekasta po}\dots\gk{ws pef'uka|men ka`i pr`os t`o ta\~uta poie\~in} \dots \\
Editor  & \gk{ponhr\~ws pef'ukamen} (``we are by nature base'');\ \ or\ \ \gk{poie\~in <ws pef'ukamen} (``to do as is in our nature'') \\
Apollo Restore  & \gk{poik'ilws pef'ukamen} (``we are by nature diverse''; top-1, \emph{\textit{plausible}}); \quad \gk{poie\~in <ws pef'ukamen} (top-2, matches editor) \\
\addlinespace[2pt]
\midrule
\multicolumn{2}{@{}l}{\textbf{Stone inscription} --- I.Kition~2059 = CIG~2645 (building inscription, Roman Cyprus)} \\
\midrule
Context & \dots\ \gk{>episkeu['asantes ...] >Is'idwros M'a[rkou ...] Seko'undou [...] u<i'os, M\~arkos ...} \gk{>an'ejhkan >e[k t\~wn >id'iwn ...] K'elson Fab[']ia(?)} \dots \\
Editor  & \gk{>an'ejhkan >e[k t\~wn >id'iwn]} (``they erected it from their own means'') \\
Apollo Restore  & \gk{>an'ejhkan >'e[tous ...]} (``they erected it in year \dots''; top-1, plausible); \quad \gk{>an'ejhkan >e[k t\~wn >id'iwn]} (top-2, matches editor) \\
\bottomrule
\end{tabularx}
\caption{Worked restoration examples across all three genres, with surrounding
context and English glosses.}
\label{tab:examples-appendix}
\end{table*}

\section{Corpus Construction Details}
\label{sec:appendix-data}

\paragraph{Corpus of Open Greek (COG).}
COG aggregates public-domain and open-access transcriptions. We began with the
Greek of the Open Greek and Latin project (OGL; 30.9M words), then identified
Greek in pre-1900 books in the Institutional Books~1.0 corpus (IB1), re-OCR'd in
2025 with improved polytonic-Greek quality~\citep{cargnelutti_institutional_2025}
(527M words after filtering). For the \emph{Patrologia Graeca} (PG), whose
two-column layout confused IB1's analysis, we used Princeton University Library
OCR with layout information (32.4M words), plus 3.7M words of miscellaneous
public-domain works. About 69\% of COG (415M of 594M words) predates 1453, the
usual cutoff for Ancient Greek; later Greek is in the conservative
\emph{katharevousa} standard.

\paragraph{Dating and deduplication.}
Because IB1 books are printed after 1450 (mostly post-1800), we estimated the date
of the \emph{Greek text} each book contains by aligning IB1 against known works
from OGL and PG: a book with $\geq 50\%$ ordered overlap was labelled an edition
of that text. Remaining books were manually sorted into pre-1453, post-1453, OGL,
and secondary (commentaries, dictionaries, grammars---excluded from COG). Slightly
more than half of COG's IB1 text (306M of 527M words) is marked as duplicated,
enabling deduplicated sampling.

\paragraph{Leiden apparatus.}
Editorial notation~\citep{van_groningen_signis_1932} is resolved consistently:
character-extent lacunae become hyphen runs of the stated length, unknown-extent
gaps become an ellipsis, and uncertain readings are handled per
Appendix~\ref{sec:appendix-uncertain}. Diacritics are stripped (final sigma
preserved, iota subscript expanded to adscript), digits removed, punctuation
collapsed to a middle-dot, and the hyphen reserved as a gap marker.

\section{Baseline Reproduction Details}
\label{sec:appendix-baselines}

We evaluate both baselines from their public releases, unmodified. Cullhed's
\texttt{Papy\_2\_Llama-3.1-8B-Instruct} is served on two GPUs; each lacuna is
presented in its instruction format with the source-appropriate system prompt and
an ``[$N$ missing letter(s)]'' marker. Ithaca is served on four GPUs in its native
format, the gap marked by exactly $N$ ``\texttt{?}'' placeholders. We change
neither model's preprocessing---both strip the sublinear uncertainty dot and keep
the underlying letter---so each sees text as its authors intended, and our test
triples are converted verbatim so all systems restore identical gaps. Two
constraints follow: Ithaca fills exactly its placeholder count (no hint-free
mode), restricts input to 50--750 characters, caps restorations at 20 characters,
and applies only to inscriptions; Cullhed requires an explicit count and has no
estimated-range mode. Apollo Restore is subject to neither restriction.

\section{Uncertain-Context Ablation}
\label{sec:appendix-uncertain}

Section~\ref{subsec:uncertain-context} adopts the uncertain-included setting
(uncertain readings shown in context) as the headline condition, matching the
baselines. Table~\ref{tab:uncertain-ablation} quantifies that choice by
re-evaluating Apollo Restore with every uncertain context character masked
(uncertain-excluded); targets and references are identical across conditions. The
effect is small and consistently positive: legible context around a gap helps
slightly, and the headline setting neither manufactures nor inflates Apollo Restore's
advantage.

\begin{table}[h]
\centering\small
\begin{tabular}{@{}lccc@{}}
\toprule
\textbf{Corpus} & \textbf{Excluded} & \textbf{Included} & \textbf{$\Delta$} \\
\midrule
DDB  & \result{0.358} & \result{0.380} & \result{$+$0.022} \\
DCLP & \result{0.134} & \result{0.143} & \result{$+$0.009} \\
IG   & \result{0.198} & \result{0.206} & \result{$+$0.008} \\
\bottomrule
\end{tabular}
\caption{Uncertain-context ablation: Apollo Restore uniform-weighted Top-1 on synthetic
gaps, uncertain-excluded vs.\ the uncertain-included headline setting,
space-agnostic.}
\label{tab:uncertain-ablation}
\end{table}

\section{Multi-Lacuna Inference}
\label{sec:appendix-multigap}

Evaluation decomposes documents into independent single-gap examples for
unambiguous scoring, but the deployed tool handles texts with several lacunae.
Gaps are filled left-to-right: for the current gap, all preceding text (with
earlier gaps already filled) is the prefix and all following text is the suffix,
with any remaining unfilled markers in the suffix replaced by an ellipsis---the
token seen in training for unknown-extent gaps. The top beam is inserted and
processing advances. Because earlier gaps typically enjoy more right-hand context,
this ordering gives a stable foundation for later gaps; joint or iterative
strategies remain future work.

\section{Additional Results}
\label{sec:appendix-extra}

\paragraph{Full metrics including legacy (unweighted).}
Table~\ref{tab:full-results} reports both unweighted (sample-mean, as in prior
work) and uniform-weighted metrics. The gap between them quantifies the short-gap
bias discussed in \S\ref{subsec:metrics}.

\begin{table}[h]
\centering\small
\setlength{\tabcolsep}{4pt}
\begin{tabular}{@{}llcccc@{}}
\toprule
& & \textbf{T1} & \textbf{T20} & \textbf{W.T1} & \textbf{W.T20} \\
\midrule
\multirow{2}{*}{DDB}  & Apollo Restore  & \result{0.399} & \result{0.589} & \result{0.432} & \result{0.632} \\
                      & Cullhed & \result{0.149} & \result{0.263} & \result{0.155} & \result{0.272} \\
\midrule
\multirow{2}{*}{DCLP} & Apollo Restore  & \result{0.188} & \result{0.350} & \result{0.175} & \result{0.327} \\
                      & Cullhed & \result{0.060} & \result{0.115} & \result{0.048} & \result{0.094} \\
\midrule
\multirow{2}{*}{IG}   & Apollo Restore  & \result{0.229} & \result{0.364} & \result{0.248} & \result{0.393} \\
                      & Ithaca  & \result{0.163} & \result{0.222} & \result{0.178} & \result{0.241} \\
\bottomrule
\end{tabular}
\caption{Full results: unweighted sample-mean over the full gap range
(T1/T20) vs.\ uniform-weighted over $1$--$20$ (W.T1/W.T20), space-agnostic,
uncertain-included, exact hint.}
\label{tab:full-results}
\end{table}

\paragraph{Per-period perplexity.}
The perplexity reduction (\S\ref{subsec:perplexity}) is consistent across
eras: classical texts fall \result{$-72.6\%$} (\result{4.89} vs.\ \result{17.83})
and post-classical \result{$-60.2\%$} (\result{5.35} vs.\ \result{13.44}), both
converging near an Apollo Restore perplexity of~\result{$5$}, indicating internalisation
of shared morphosyntactic structure across Greek's temporal variants.

\section{Reproducibility and Training Details}
\label{sec:appendix-repro}

\paragraph{Base architecture.}
Apollo Restore fine-tunes Mistral Small Base 2503 \citep{jiang2023mistral7b}, a
decoder-only Transformer \citep{vaswani2017attention} with $40$
layers, model dimension $5{,}120$, feed-forward dimension $32{,}768$, $32$
attention heads and $8$ key/value heads (grouped-query attention;
\citealp{ainslie2023gqa}), head dimension $128$, RMSNorm, rotary position
embeddings \citep{su2024roformer} ($\theta{=}10^{8}$), and a $131{,}072$-token
byte-pair \citep{sennrich2016bpe} Tekken tokenizer. We do not extend the
vocabulary: the FIM control tokens are reserved tokens already present.
Long-context RoPE scaling (YaRN) is disabled for the $4{,}096$-token window.
Table~\ref{tab:repro-hparams} lists the complete training configuration.

\begin{table}[h]
\centering\small
\begin{tabular}{@{}lp{4cm}@{}}
\toprule
\textbf{Hyperparameter} & \textbf{Value} \\
\midrule
Optimiser & AdamW~\citep{loshchilov2019adamw} $\beta{=}(0.9,0.95)$, $\varepsilon{=}10^{-8}$ \\
Weight decay & $0.1$ \\
LR schedule & cosine; peak $1\times10^{-5}$, floor $1\%$ \\
Warmup & $1{,}500$ steps (linear) \\
Total steps & $10{,}000$ \\
Gradient clipping & global-norm $1.0$ \\
Sequence length & $4{,}096$ tokens \\
Global batch & $65{,}536$ tokens ($8{\times}2{\times}4{,}096$) \\
Precision & bf16 compute / fp32 master \\
Parallelism & FSDP2/HSDP (shard $8$), TP${=}$CP${=}1$ \\
Activation checkpointing & full \\
Checkpoint / val interval & $2{,}000$ / $500$ steps \\
Hardware & $1$ node $\times\,8$ H100 \\
\bottomrule
\end{tabular}
\caption{Complete training hyperparameters for the Apollo Restore model.}
\label{tab:repro-hparams}
\end{table}

\paragraph{Software and serving.}
Training uses Mistral's \emph{Forge} with fully sharded data parallelism
\citep[FSDP2/HSDP;][]{zhao2023pytorchfsdp} and streaming Mosaic Data Shard
(MDS) loading; sources are mixed by normalised sampling proportions and FIM
samples packed to fill the window. Runs launch on Slurm via \texttt{srun}. Key
dependencies: PyTorch~$2.10$, xFormers~$0.0.35$, MosaicML \texttt{streaming}~$0.12$.
Checkpoints are written in PyTorch Distributed Checkpoint (DCP) format,
consolidated to a bf16 \texttt{safetensors} file, and served with
vLLM~\citep{kwonEfficientMemoryManagement2023} (beam search, 20 candidates).

\section{Ablation Study}
\label{sec:appendix-ablation}

The final model is the product of systematic experimentation across four pipeline
axes (Table~\ref{tab:sweep-results}), all at $10{,}000$ steps from a shared
initialisation with architecture and schedule fixed. The initial model (v0) used
a Gamma gap-length distribution skewed to short gaps, no hints, and a raw-volume
mix (DDB only 2.1\% of tokens), giving weighted Top-1 \result{0.292} on DDB with
severe under-generation ($\Delta_{\mathrm{len}}=-6.0$).

\begin{table}[h]
\centering\small
\begin{tabular}{@{}lccc@{}}
\toprule
\textbf{Configuration} & \textbf{W.T1} & \textbf{W.T20} & $\boldsymbol{\Delta_{\mathrm{len}}}$ \\
\midrule
Hints $+$ uniform $+$ 3$\times$DDB  & \textbf{\result{0.456}} & \textbf{\result{0.647}} & \result{$-$0.9} \\
Uniform $+$ 3$\times$DDB (no hint)  & \result{0.388} & \result{0.669} & \result{$-$3.6} \\
3$\times$DDB only (no hint)         & \result{0.356} & \result{0.643} & \result{$-$5.2} \\
Uniform only (no hint)              & \result{0.326} & \result{0.613} & \result{$-$5.3} \\
v0 (baseline)                       & \result{0.292} & \result{0.584} & \result{$-$6.0} \\
\bottomrule
\end{tabular}
\caption{Ablation sweep on 2{,}000 held-out DDB samples. Length-hint conditioning
is the largest single improvement; uniform sampling and DDB upsampling each help,
and their combination yields the final configuration.}
\label{tab:sweep-results}
\end{table}

\paragraph{Negative results.}
Two directions we expected to help did not. \emph{Modern-Greek continued
pre-training} (90{,}000 steps, 16 GPUs) reached only weighted Top-1 \result{0.136}
on DDB---below v0---because the learning rate needed for any signal overwrote the
CPT representations; all final models fine-tune directly from the base.
\emph{Diachronic curriculum} training (broad prose, then specialising onto
fragmentary sources) overfit the small fragmentary corpora and degraded held-out
accuracy relative to a single-stage mixture, so the final model trains in one
stage.

\end{document}